\documentclass[runningheads]{llncs}
\usepackage[T1]{fontenc}
\usepackage{graphicx}
\usepackage{amsmath}
\usepackage{amssymb}
\usepackage{booktabs}
\usepackage{gensymb}
\usepackage{tabu}

\usepackage{svg}
\usepackage[pagebackref,breaklinks,colorlinks]{hyperref}
\usepackage[capitalize]{cleveref}
\newcommand{\etal}{\textit{et al.}}

\begin{document}
\title{Reappraising Domain Generalization \\ in Neural Networks}
\titlerunning{reappraising DG for NN}

\author{%
Sarath Sivaprasad $^{1,2}$ \mbox{ } Akshay Goindani $^{1}$ \mbox{ }  Vaibhav Garg$^{1}$ \mbox{ } \\ Ritam Basu $^{1}$ \mbox{ }  Saiteja Kosgi $^{1}$  \mbox{ } Vineet Gandhi$^{1}$ \\ \vspace{5mm}
 $^{1}$Kohli Centre on Intelligent Systems, IIIT Hyderabad \mbox{ } $^{2}$TCS Research, Pune \mbox{ } \\ 
 \small{\texttt{\{sarath.s, akshay.goindani, saiteja.k, ritam.basu\}@research.iiit.ac.in}} \mbox{  }\\
\small{\texttt{vaibhav.garg@students.iiit.ac.in, vgandhi@iiit.ac.in}} \\
}

\maketitle

\begin{abstract}


Given that Neural Networks generalize \emph{unreasonably} well in the IID setting (with benign overfitting and betterment in performance with more parameters), OOD presents a consistent failure case to better the understanding of how they learn. This paper focuses on Domain Generalization (DG), which is perceived as the front face of OOD generalization. We find that the presence of multiple domains incentivizes domain agnostic learning and is the primary reason for generalization in Tradition DG. We show that the state-of-the-art results can be obtained by borrowing ideas from IID generalization and the DG tailored methods fail to add any performance gains. Furthermore, we perform explorations beyond the Traditional DG (TDG) formulation and propose a novel ClassWise DG (CWDG) benchmark, where for each class, we randomly select one of the domains and keep it aside for testing. Despite being exposed to all domains during training, CWDG is more challenging than TDG evaluation. We propose a novel iterative domain feature masking approach, achieving state-of-the-art results on the CWDG benchmark. Overall, while explaining these observations, our work furthers insights into the learning mechanisms of neural networks.
\end{abstract}

\section{Introduction}
\label{sec:intro}
Generalization is a key goal in machine learning. Empirical results show that sufficiently parameterized networks can completely fit any training data in any configuration of labels~\cite{zhang_fittrain}. Hence, the most rudimentary notion of evaluation is to measure performance on unseen Independent and Identically Distributed (IID) data. This notion is brittle for real-world settings; for instance, minimal perturbations can substantially deteriorate the performance of models trained in IID setting~\cite{hendrycks2018benchmarking}. Significant efforts have been made to improve the generalization of the neural network across perturbations. However, more and more evidence is arising that robustness from synthetic image perturbations like noise, simulated weather artifacts, adversarial examples, etc., does not necessarily improve performance on distribution shift arising in real data~\cite{taori2020measuring}. Furthermore, performance on IID test data does not necessarily imply that the network has learned the expected underlying distributions~\cite{geirhos2020shortcut}. Hence, evaluating generalization across Out of Distribution Data (OOD) is important for machine learning models.

\begin{figure*}[t]
  \centering
   \includegraphics[width=\linewidth]{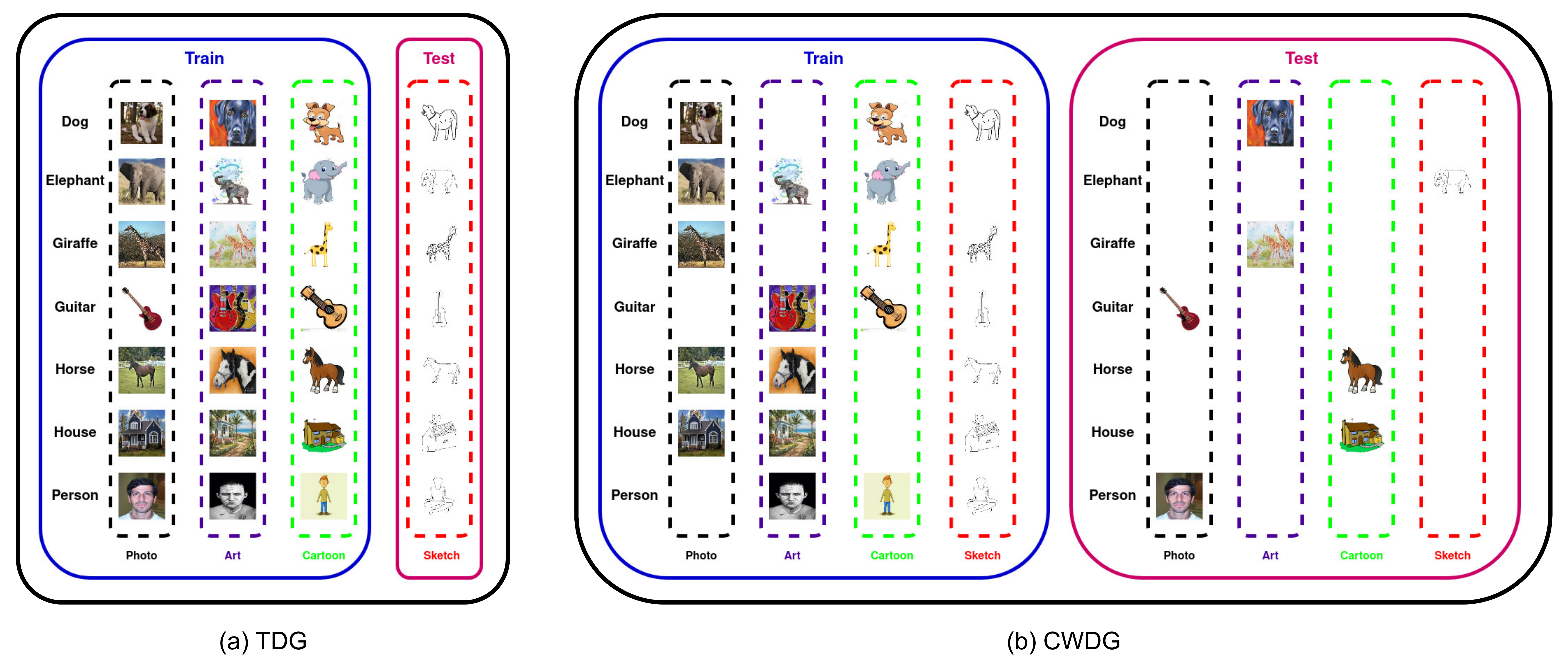}
   \caption{The figure shows the difference in train test split with Traditional DG (TDG) setting and ClassWise-DG (CWDG) setting on PACS dataset~\cite{PACS}. (a) shows one of the four splits in TDG, (b) shows one train test split out of the 16384 possible samplings. The open entry in the train set of CWDG corresponds to the entry in the test set. 
   }
   \label{fig:teaser}
\end{figure*}

Domain Generalization (DG) goes beyond perturbations and is a common way to formally evaluate OOD generalization. It focuses on scenarios where target domains have \emph{distinct characteristics} but \emph{no data for training}. DG aims to extract domain-agnostic features by learning from multiple domains, which can then be applied to an unseen domain. For instance, learn a model using labeled data for discriminating classes in photos, paintings, cartoons and then apply it on sketches (traditional DG setup is illustrated in Figure~\ref{fig:teaser}(a)). The motivation of DG is to produce human-like classification/representation models with deeper semantic sharing across domains - a horse is a horse irrespective of its form of depiction (a photo, cartoon, painting, or a sketch).

A myriad of \emph{inventive methods} has been proposed for Traditional Domain Generalization (TDG). Some of the notable efforts include reverse gradients to obtain domain agnostic features~\cite{ganin2015unsupervised}; kernel methods~\cite{muandet2013domain} to learn to map all domains to a common representation; style transfer for data augmentation~\cite{stylization}; jointly training domain agnostic and domain-specific models~\cite{Undo_bias} and inhibiting features corresponding to the highest gradients in each domain~\cite{RSC}. Wang et al.~\cite{wang2021generalizing} presents a comprehensive review of over 150 methods. Recently, Gulrajani and Lopez-Paz~\cite{gulrajani2020search} show that an Empirical Risk Minimization (ERM) baseline gives a competent performance on TDG benchmarks, and none of the tailored methods give \emph{any clear and consistent advantage} over the baseline. The findings in~\cite{gulrajani2020search} motivate the community towards reappraisal of DG and a more fundamental analysis around the OOD generalization in Neural Networks.

Our paper makes effort in this direction. We perform elemental experiments by training on a single domain and testing all other domains. We then augment the training data by successively adding one domain at a time until all but the test domain is in the train data. Our experiments suggest that in the presence of multiple domains, the networks tend to fit the most low lying variance, i.e., discriminative features shared across domains. Hence, learning the domain agnostic features is most convenient for the network and not a challenge, explaining why ERM already gives state-of-the-art (SOTA) results in TDG. We show that the performance of ERM in TDG can be fully explained using the distribution shift between train and test data. We further demonstrate that factors that aid IID generalization (optimization, augmentation, backbone) play a key role in TDG. We find that the current DG tailored methods add no value over an optimally trained ERM baseline. We observe that generalization properties of SGD~\cite{Chaudhari} hold firm in TDG, and the robustness of a backbone in DG is proportional to its IID generalization. Our analysis helps achieve a new SOTA on six different DG datasets. 

Our experiments motivate us to go beyond TDG and explore scenarios where fitting the low lying variance is detrimental to generalization. To this end, we propose an alternate setting to evaluate the generalization of neural networks: ClassWise-DG (CWDG). In this setting, a randomly selected domain from each class is kept aside for testing, as illustrated in Figure~\ref{fig:teaser}(b). Overall the model sees all domains during training, but only a subset of domains are seen for each class. We argue that CWDG benchmarking shares similarities with human learning. For example, a child in the early years may see some objects in photos and sketches, some in paintings and cartoons; but may not necessarily see all classes in all domains. In contrast, the TDG setting assumes that all class categories are seen in all source domains, and the target domain is never seen (suddenly, you are exposed to sketches one day!). Moreover, in the real world, the availability of class annotations is not uniform across domains; therefore, performance on the CWDG benchmark will show the model's efficacy in leveraging different kinds of supervision available in the real world. Contrary to the intuitive expectation, we observe that neural networks struggle to retain their performance despite seeing all the domains during training.

We argue that CWDG creates an incentive to learn domain-specific features and makes generalization difficult. In Figure~\ref{fig:synth} we show that learning the low lying variance in CWDG can lead to shortcut learning~\cite{geirhos2020shortcut}. Learning to discriminate among domains creates a prior in CWDG, and hence learning domain-agnostic features becomes difficult. In contrast, in TDG, since all classes are seen in similar proportion across domains, learning domain-agnostic features is convenient for the network. We believe CWDG emulates domains shifts present in real-world datasets and is a more robust benchmark to evaluate the ability to learn domain-agnostic features. To address the challenges in CWDG, we propose a method to suppress mask domain specific features. The proposed method gives an average of 8\% improvement over ERM on the PACS dataset. Overall, our work makes the following contributions:

\begin{itemize}
    \item We present empirical experiments which help explain the findings in~\cite{gulrajani2020search} and uncovers \emph{why ERM works}. We also present ablations studies over IID generalization methods, leading to new SOTA on six DG dataset. 
     \item Our analysis shows that the challenge of OOD generalization may lie beyond TDG formulation. The distribution shift across classes appears as the primary hurdle. To this end, we propose the novel CWDG formulation.
    \item We perform thorough benchmarking on CWDG using popular algorithms and propose a novel method surpassing the SOTA by a significant margin. 
 
\end{itemize}

\section{Related Work}
\label{sec:related work}

We discuss the prior art in two components. We first address the common notion of generalization in IID data. Subsequently, we discuss the previous works on TDG and motivate the need for the new CWDG setting.

\paragraph{Generalization in IID setting:} Sufficiently parameterized networks can completely fit any training data~\cite{zhang_fittrain}. Hence, a bare essential way to evaluate a neural network is to train on a randomly selected portion of the data and test on the unseen part. Popular methods like dropout~\cite{srivastava2014dropout}, weight decay, early stopping, and regularization techniques~\cite{wan2013regularization,gastaldi2017shake} have shown to improve this notion of generalization. It is common wisdom that spatial transforms in image data help improve generalization. Constraining networks~\cite{bach2017breaking,sivaprasad2020curious} have also shown to improve generalization in IID setting. The optimizer also plays a role in generalization; specifically, SGD is shown to achieve better generalization than adaptive algorithms~\cite{wilson2017marginal}. 

Recently Taori~\etal~\cite{taori2020measuring} suggests considering generalization beyond IID setting with perturbations. They report a thorough study with 204 ImageNet models, showing that robustness from synthetic image perturbations like noise, simulated weather artifacts, adversarial examples, etc., does not improve the performance on distribution shift arising in real-world data. Moreover, Recht~\etal~\cite{recht2018cifar,recht2019imagenet} expose the problems in using a specific part of IID distribution as test data. They show a drop in performance when tested on new test data collected from the same distribution, motivating the evaluation beyond the IID setting.

\paragraph{Domain generalization:} 

Our work focuses on the DG in deep neural networks, and for pre-deep learning efforts, we refer the reader to the review by Moreno~\etal~\cite{moreno2012unifying}. Furthermore, we limit our discussion to DG in image classification. TDG formulation involves learning from multiple domains and testing on an unseen domain. TDG on image classification is commonly evaluated on six datasets: \cite{PACS,fang2013unbiased,officehomevenkateswara2017deep,IRMarjovsky2019invariant,domainnetpeng2019moment,rmnistghifary2015domain}. The details of these datasets are illustrated in Table~\ref{Tab:Datastats}. We perform experiments on all these datasets.

Learning domain agnostic features using the TDG formulation has seen significant interest in recent years. The problem has been approached from many different directions like data augmentation~\cite{stylization,zhou2021domain,MIXUPxu2020adversarial}, gradient manipulation~\cite{ganin2015unsupervised,RSC}, ensemble learning~\cite{mancini2018best}, and feature disentanglement~\cite{Undo_bias,piratla2020efficient}. For a comprehensive list, readers can refer to the recent surveys~\cite{wang2021generalizing,zhou2021domain_survey}. It is worth noting that several of these ideas~\cite{ganin2015unsupervised} have found widespread success beyond the TDG setting. 

Gulrajani and Lopez-paz~\cite{gulrajani2020search} suggest that inconsistencies in experimental conditions (datasets and training protocols) render fair comparisons difficult. They propose DomainBed, a unifying benchmark for TDG. They empirically show that a carefully implemented ERM outperforms the state-of-the-art in terms of average performance. A natural thought that arises is that despite the proposition of numerous inventive ideas for TDG, why none of them improves over the baseline ERM. In this work, we claim that TDG is not an appropriate formulation to measure the efficacy of a model to learn domain agnostic features, at least in the current form. We argue that in TDG formulation, learning domain agnostic features is the most convenient thing for the network, not a challenge. 

Consequently, we propose CWDG, a more challenging DG formulation, which leaves room for shortcut learning~\cite{geirhos2020shortcut}. Our work interestingly contrasts with~\cite{maniyar2020zero} which proposes further \emph{constraints} on TDG by introducing unseen classes in the test domain. We instead \emph{relax} the assumptions and expose all domains during training. 

\begin{table*}[t]
\begin{center}
\scriptsize
\begin{tabular}{| l | c | c | c | c |}
\hline
 \textbf{Dataset}& \textbf{\# D} & \textbf{Domains} & \textbf{\# C} & \textbf{\# Images}\\
\hline
RMNIST~\cite{rmnistghifary2015domain} & 6 & 0\degree, 15\degree, 30\degree, 45\degree, 60\degree, 75\degree & 10 & 70000 \\ \hline
CMNIST~\cite{IRMarjovsky2019invariant} & 2 & Red, Green & 2 & 120000 \\ \hline
DomainNet~\cite{domainnetpeng2019moment} & 6 & Clipart, Infograph, Painting, Quickdraw, Real, Sketch & 345 & 586575 \\ \hline
PACS~\cite{PACS} & 4 & Photo, Art-Painting, Cartoon, Sketch &  7 & 9991\\ \hline
VLCS~\cite{fang2013unbiased} & 4 & Caltech101, LabelMe, SUN09,  VOC2007 & 5 & 10729 \\ \hline
Office-Home~\cite{officehomevenkateswara2017deep} & 4 & Art, Clipart, Product, Photo & 65 & 15558 \\ \hline

\end{tabular}
\end{center}
\caption{The table presents the statistics of datasets we use for evaluation.}
\label{Tab:Datastats}
\end{table*}

\section{Why ERM works in TDG?}

\begin{figure}[t]
    \centering
       \includegraphics[width=0.68\linewidth]{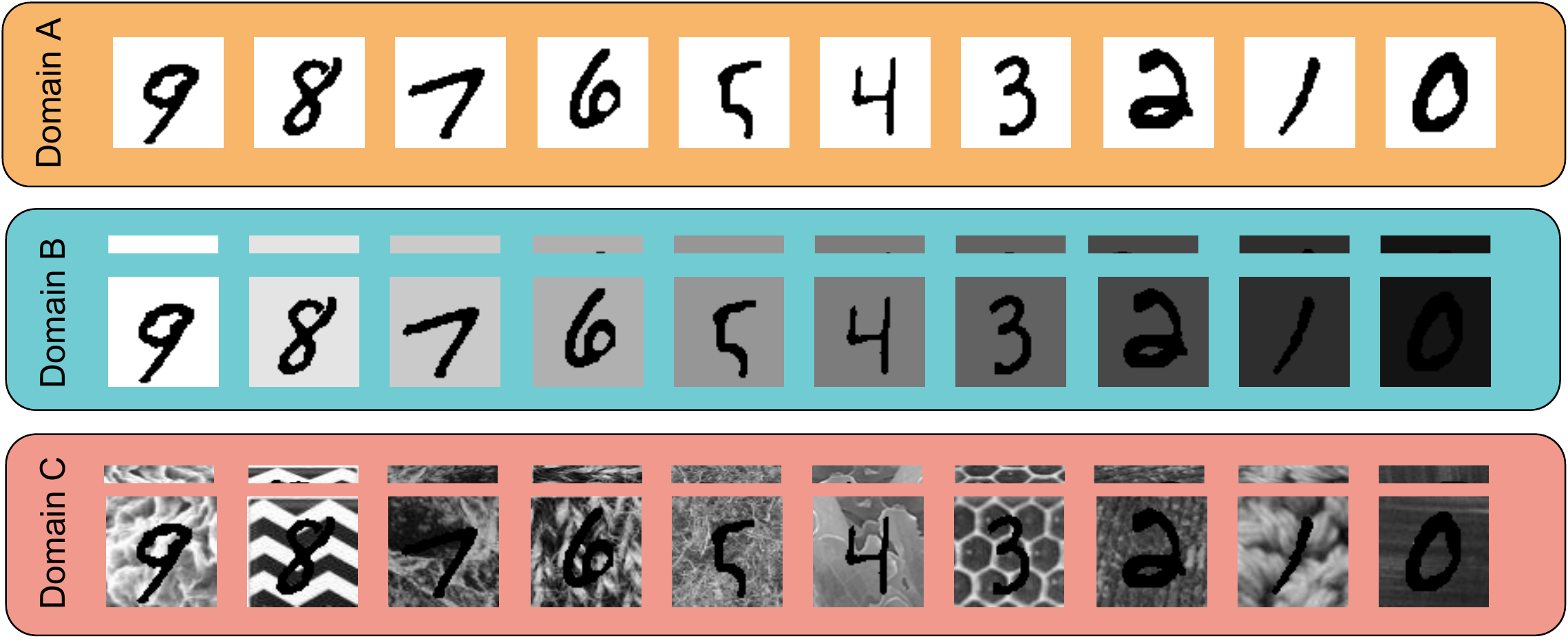} \includegraphics[width=0.31\linewidth]{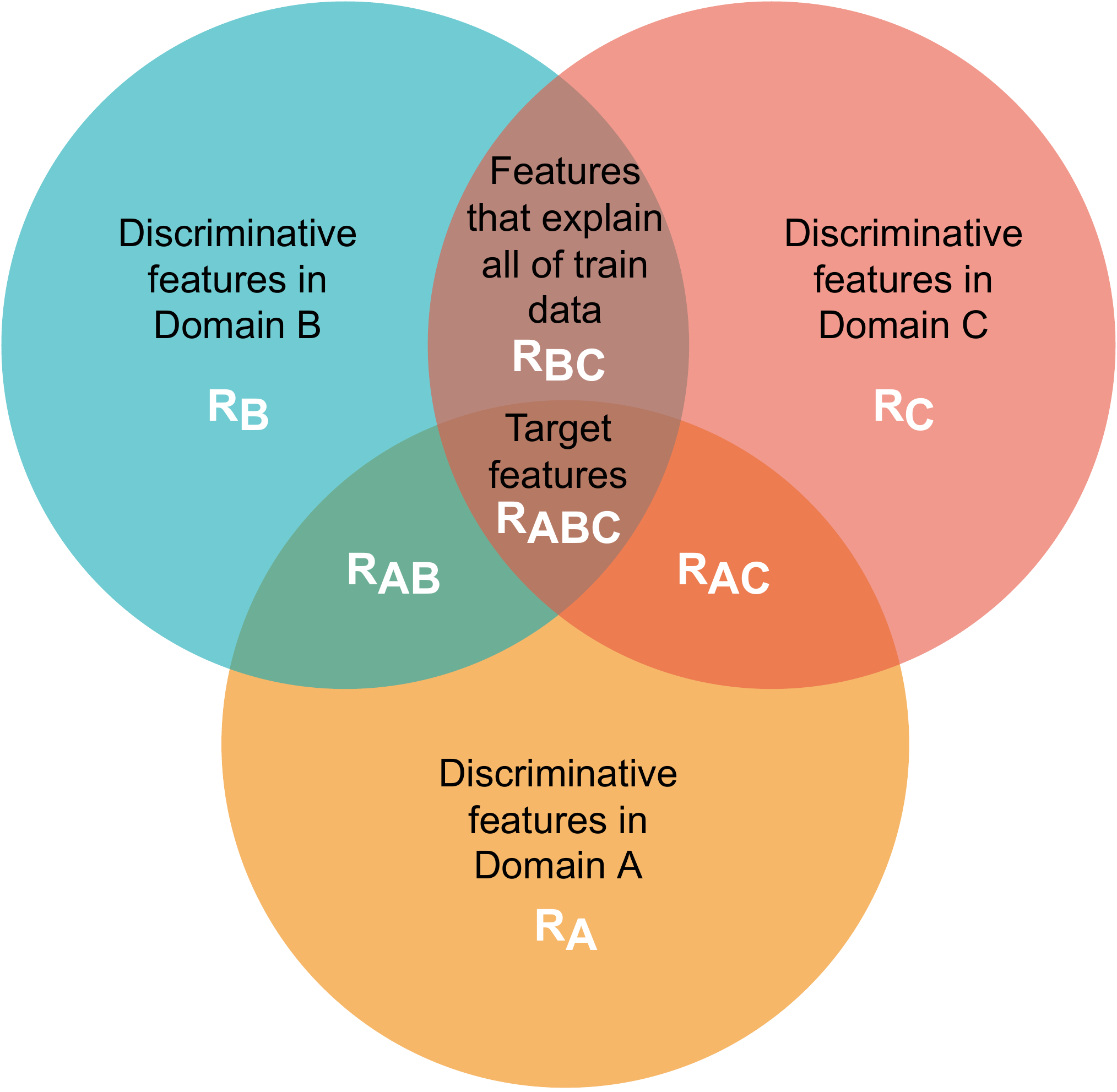}  
    \caption{ Left: MNIST data in three different domains. Right: $R_{BC}$ is the low lying variance that explain the train data when jointly trained on domains $B$ and $C$. Restricting the feature space to $R_{BC}$ will improve performance when tested on domain $A$.}
    \label{fig:DG_venn}
\end{figure}

We hypothesize that ERM works so well in TDG because networks tend to learn the low lying variance first. We back this hypothesis with empirical results. Consider the simple classification setup in Figure~\ref{fig:DG_venn}(a), where a three-layer CNN is trained to classify among the ten digits. The data is presented in three domains, the first is normal MNIST data (domain A), the second represents each digit with a different grays-shade in the background (domain B), and the third represents each digit with a different texture in the background (domain C). We train models using domains B and C individually and test on A; the models lead to 16.4\% and 14.8\% accuracy, respectively. The network learns shortcuts (low-lying variance), i.e., the shade or the texture, and not the desired shape of the digit. Training together with domain B and domain C and testing on domain A leads to 24\% accuracy. We argue that the network tends to again learn the low-lying variance i.e., the intersection of the feature spaces of domains B and C. 

 Assume $R_A$, $R_B$ and $R_C$ be the span of learnt features corresponding to domain $A$, $B$ and $C$ respectively, when trained on the digit classification task (Figure~\ref{fig:DG_venn}(b)). Assume $R_{BC}$ be the intersection of $R_B$ and $R_C$. When the model is trained jointly on domains B and C, $R_{BC}$ corresponds to the low lying variance that explains the entire train data. Learning the low lying variance here is guaranteed to improve domain generalization ($\frac{P(R_A \cap R_B \cap R_C)}{P(R_B \cap R_C)} \ge \frac{P(R_A \cap R_B \cap R_C)}{P(R_B)}$ and $\frac{P(R_A \cap R_B \cap R_C)}{P(R_B \cap R_C)} \ge \frac{P(R_A \cap R_B \cap R_C)}{P(R_C)}$), possibly explaining the significant performance gains in this particular experiment. The illustration also explains why the validation is non-trivial in TDG setup, where even perfectly restricting the feature space to $R_{BC}$ is not enough to guarantee performance on domain $A$. Furthermore, the analysis suggests that adding domains to the train data should improve performance in TDG. The next section presents experiments on the PACS dataset for further empirical validation of our hypothesis.
 
 \subsection{PACS vs PACS}
\label{sec:PACS_vs_PACS}

As the first elemental experiment, we train models on a single domain and test them on other domains. The task is to classify the seven classes in the PACS dataset. The training involves fine-tuning the Inception-Resnet~\cite{inception_resnet} backbone pre-trained on ImageNet. The model backbone is trained four times, independently on each of the four domains. We keep out 10\% of the training domain for model selection and evaluation. We evaluate each model on the kept out portion of the train domain and the entirety of the three other domains. We then repeat the training and inference procedure using a model with a two-node bottleneck layer before the classification head. In total we train eight models (four with bottleneck layer and four models without it).

\begin{figure}[t]
  \centering
  \includegraphics[width=0.7\linewidth]{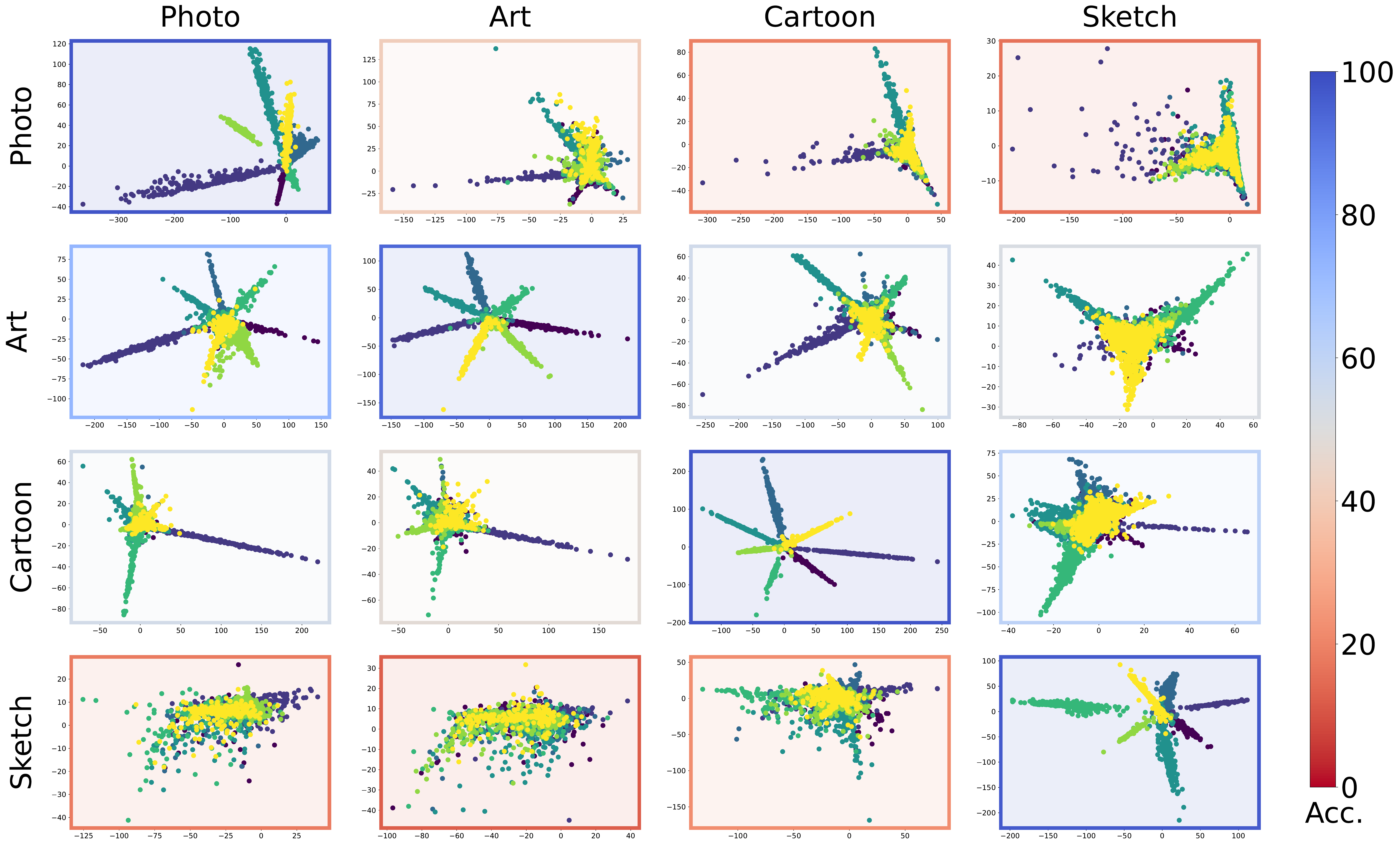}
  \caption{ The color of each tile shows the accuracy of the model trained on the domain corresponding to the row name and tested on the domain corresponding to the column name. The plot within each tile, shows the embeddings learned using an additional network with a two-node bottleneck. Embeddings from different classes are represented with different colors.}
  \label{fig:Pairwise}
\end{figure}

Figure~\ref{fig:Pairwise} shows the accuracy of the model and visualizes the features of the corresponding bottleneck models. The row and column names correspond to the train and test domain. The accuracy of the non-bottleneck model is indicated as the color of each tile in Figure~\ref{fig:Pairwise}. The features learned at the bottleneck layer are plotted within the tile, and a different color is used to indicate each class. 

Expectedly, the diagonal plots in Figure~\ref{fig:Pairwise} show that the features learned from a domain are most effective in the same domain (accuracy is high and the features among different classes are well discriminated). We can also see that features learned from a domain like ‘Art-Painting’ help discriminate samples of almost all four domains. However, features learned from the ‘Sketch’ domain are not adequate for discriminating samples in any other domain. The key takeaway from the experiments is that discriminative features from a domain do not necessarily transfer well to another domain. 

\subsection{Domain Augmentation}
\label{domain_aug}

\begin{figure}[t]
    \centering
    \begin{tabular}[t]{ccc}
        \includegraphics[height=0.32\linewidth]{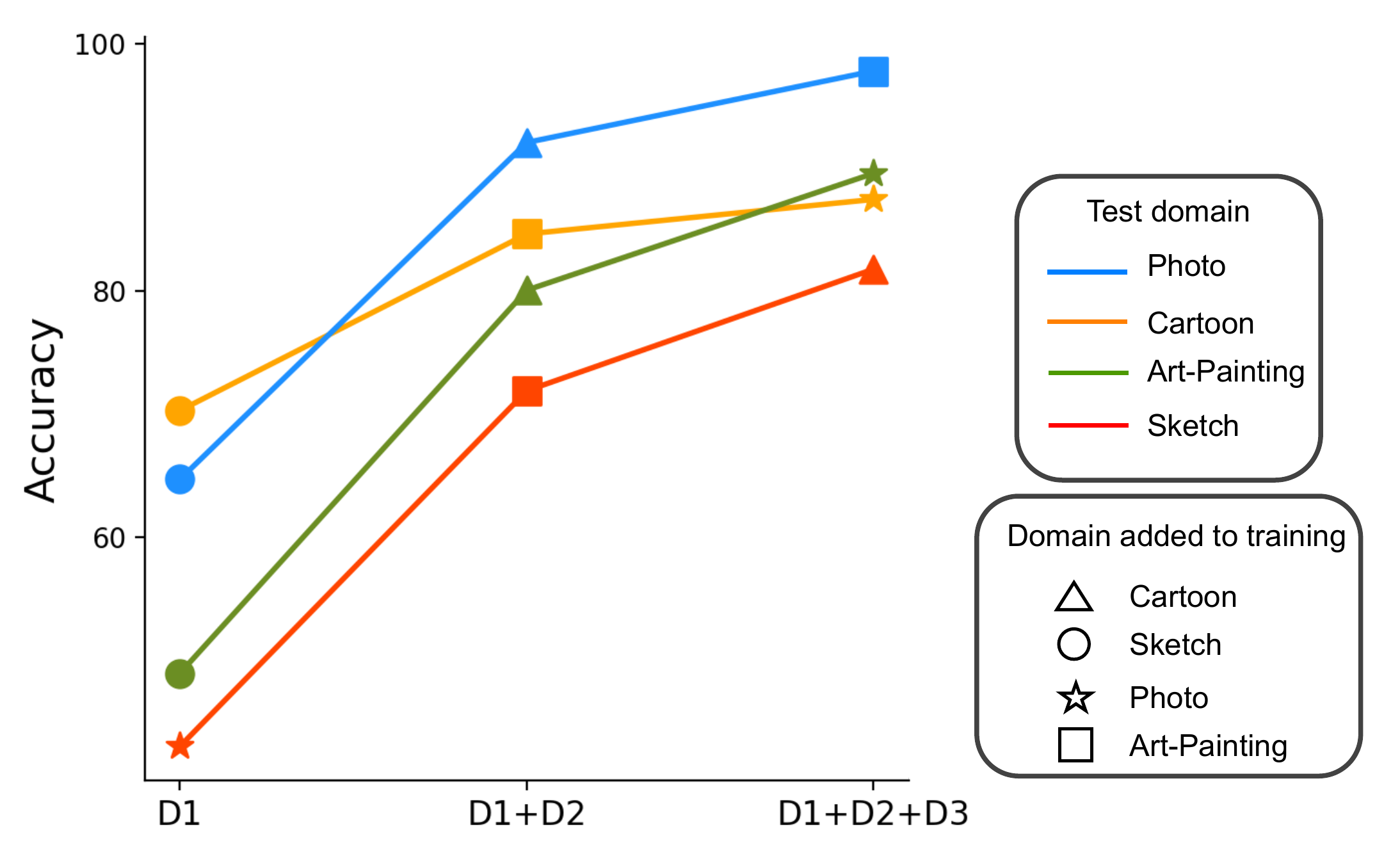}&  \includegraphics[height=0.32\linewidth]{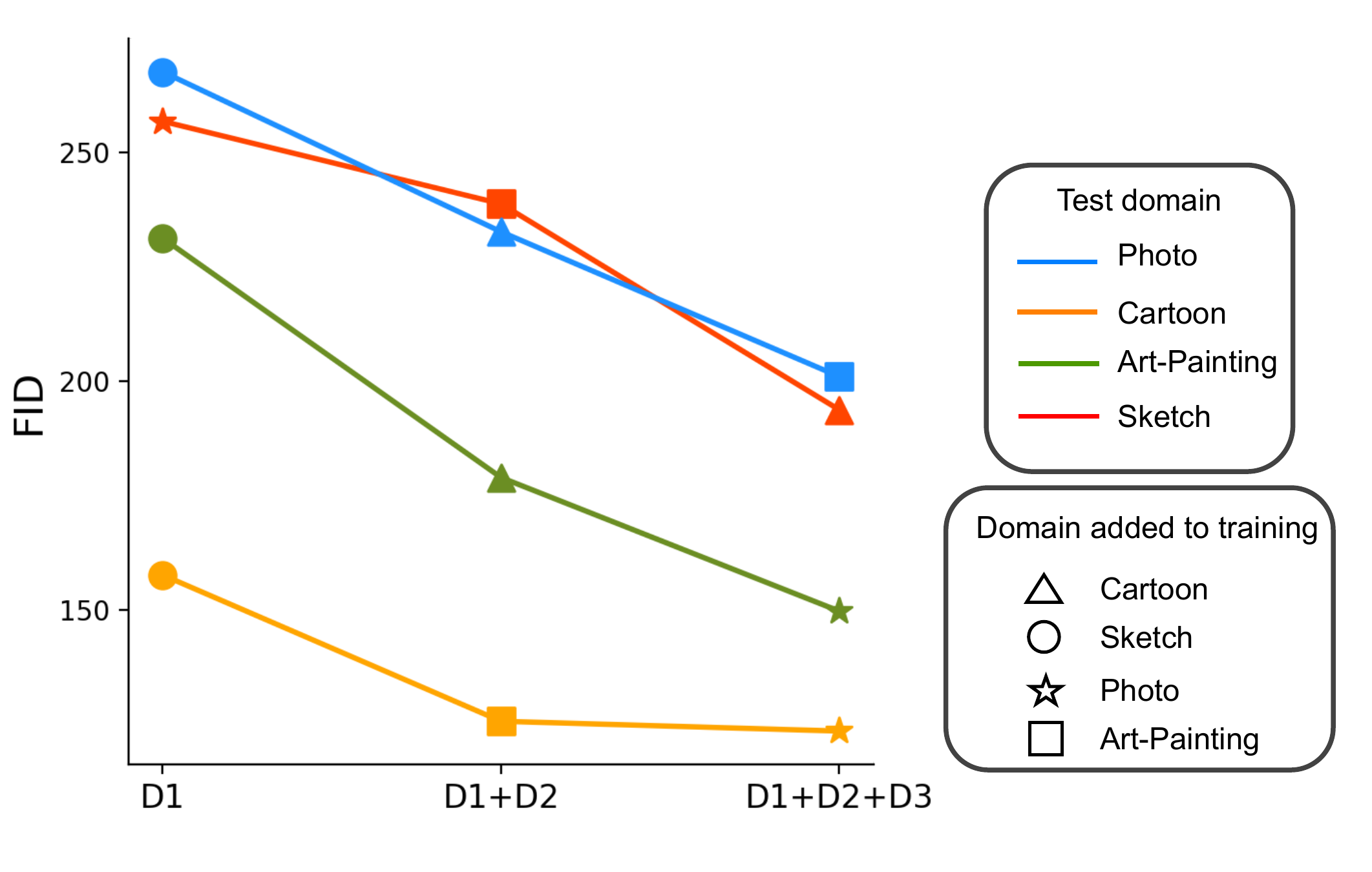}& \vspace{-1em}  \\ 
        {\scriptsize (a)} & {\scriptsize (b)} \\
    \end{tabular}
    \caption{Test performance of Inception-Resnet backbone trained on different subsets of PACS dataset. The line color denotes the domain kept out for testing. At every tick on X-axis a new domain is added to the train data. For instance, when the test domain is `Cartoon'(yellow line), the model is first trained on just `Sketch'(circle). In subsequent steps `Art-painting'(square) and `Photo'(star) are added to training data. (a) Test performance  increases on adding domains to train data and (b) The distribution shift decreases between train and test data in each subsequent step.}
    \label{fig:mulit_domain}
\end{figure}

We perform domain augmentation experiments to empirically validate the hypothesis that adding domains to train data in ERM improves performance in TDG. We start with a single domain each in the train and test splits and subsequently keep adding the remaining domains to the train set (until there are all but the test domain in the train data). For instance, we start with the `Sketch' domain as test and `Photo' as train. We add `Art-Painting' and `Cartoon' to the train data in the subsequent two steps. We compare test accuracy (on `Sketch') between the three runs. To make the comparison unbiased of the train data size, we randomly sub-sample after adding each domain such that the size of train data is the same for all experiments across iterations. We perform this experiment with all four domains as test data with a randomly selected order of adding domains. 

Figure~\ref{fig:mulit_domain}(a) presents the obtained results. Despite some domains being poor at contributing discriminative features to other domains (as seen in Section~\ref{sec:PACS_vs_PACS}), we find a significant advantage of having multiple domains in the training set. The results clearly show that adding domains to the train data, however different from the test domain, aid domain agnostic feature learning and improves test performance. This observation follows the argument presented in Figure~\ref{fig:DG_venn}(b), that adding domains creates an incentive for the ERM to learn the features at the intersection of the train domains, improving the test performance.  

We use Frechet Inception Distance (FID)~\cite{heusel2017gans} as a measure of distribution shift between the train and test split of the datasets. FID measures the distance between two multivariate Gaussians $d$ as, 

\small

$
\begin{array}{l}
d\left(\left(m_{1}, C_{1}\right),\left(m_{2}, C_{2}\right)\right)=$$ 
$$\Big(\left\|m_{1}-m_{2}\right\|_{2}^{2}+\operatorname{Tr}\left(C_{1}+C_{2}-2\left(C_{1} C_{2}\right)^{\frac{1}{2}}\right)\Big)^{\frac{1}{2}} 
\end{array}
$
\label{eq:3}

\vspace{3mm}

 \normalsize 
where $d\left(\left(m_{1}, C_{1}\right),\left(m_{2}, C_{2}\right)\right)$ is the Frechet distance, $m_1$, $m_2$ are the means, $C_1$ and $C_2$ are the covariances of two gaussian distributions, $Tr$ denotes the trace of a matrix. 

We compute the distance $d$ between the embeddings of the final layer of an InceptionV3 network~\cite{szegedy2016rethinking} pre-trained on ImageNet. To compute the FID score in each experimental setting, we pass the train and test split through the pre-trained network and calculate the distance $d$ of the embeddings. Comparing Figure~\ref{fig:mulit_domain}(a) with Figure~\ref{fig:mulit_domain}(b), we can see that adding a new domain to the train data decreases the FID distance between the train and test splits and increases the test accuracy. The experiment suggests that in the TDG formulation, the accuracy of the model is entirely explained by the distribution shift between the train and test splits (Figure~\ref{fig:mulit_domain}).

\section{Exploiting IID generalization methods for TDG}
\label{sec:TDG}

In this section, we show that methods that help IID generalization are also key while training an ERM in TDG. We demonstrate that exploiting these IID tricks leads to SOTA performance on a vanilla ERM. Following prior art in TDG, we keep aside one domain for testing in each fold and train on the other three. We use an oracle for model selection. To measure the effect of each intervention in the ablation, we keep all the other modeling choices the same and run the experiment five times and report the mean value. Unless specifically mentioned otherwise, the training protocol and hyper-parameters are the same as in DomainBed~\cite{gulrajani2020search}. The data augmentation is also the same as in DomainBed. 

The ablation experiments are limited to PACS~\cite{PACS} dataset. We train the ERM with six different backbones: AlexNet~\cite{AlexNet},  VGG-19~\cite{vgg19}, ResNet-18~\cite{resnet18}, ResNet-50~\cite{resnet50}, DenseNet-121~\cite{huang2017densely} and Inception-Resnet~\cite{inception_resnet}. We run all four folds of PACS on these backbones with SGD and ADAM optimizer. As the next intervention, we run all the aforementioned backbones with and without augmentations with an SGD optimizer. The list of augmentations and other training details are given in supplementary material.

We compare the optimized ERM baseline against the top-performing methods on DomainBed, on six different datasets (Table~\ref{Tab:Datastats}). Outside of DomainBed, we also use the multi-branch reverse-gradient (GRL)~\cite{ganin2015unsupervised} model on the Inception-ResNet backbone. We present the results of other methods in their original form. We also train all the methods on top of the best performing ERM baseline (Inception-Resnet backbone) and present the results in the supplementary material.


\begin{table*}[t]
\begin{center}
\resizebox{\textwidth}{!}{%
\begin{tabu}{|c|c|c|c|c|[2pt]c|c|c|c|[2pt]c|c|c|c|}
\hline
 & \multicolumn{4}{ c |[2pt]}{\textbf{Adam with augmentation}}  & \multicolumn{4}{ c |[2pt]}{\textbf{SGD without augmentation}} & \multicolumn{4}{ c |}{\textbf{SGD with augmentation }}\\ 
\hline
\textbf{Backbones}& \textbf{Photo} & \textbf{Sketch} & \textbf{Art} & \textbf{Cartoon} & \textbf{Photo} & \textbf{Sketch} & \textbf{Art} & \textbf{Cartoon} & \textbf{Photo} & \textbf{Sketch} & \textbf{Art} & \textbf{Cartoon} \\
\hline

Alexnet  	&	78.05	&	58.72	&	60.56	&	64.13	&	88.26	&	60.42	&	65.645	&	70.065	&	87.69	&	69.17	&	66.91	&	69.28	\\	\hline
Vgg-19\_BN	&	80.63	&	69.62	&	69.08	&	70.5	&	88.41	&	77.63	&	76.31	&	70.47	&	84.27	&	82.12	&	70.65	&	79.6	\\	\hline
Resnet-18	&	83.21	&	67.83	&	69.47	&	76.07	&	87.96	&	74.38	&	75.7	&	77.38	&	87.34	&	80.29	&	73.64	&	76.91	\\	\hline
Resnet-50	&	83.08	&	70.85	&	65.31	&	76.97	&	88.53	&	78.21	&	72.99	&	79.28	&	87.57	&	81.02	&	74.91	&	77.72	\\	\hline
DenseNet-121	&	86.21	&	68.72	&	71.64	&	74.81	&	87.5	&	79.1	&	73.4	&	76.44	&	88.9	&	80.42	&	74.31	&	78.15	\\	\hline
Inc-Resnet	&	89.27	&	71.28	&	73.4	&	76.81	&	96.12	&	81.36	&	84.96	&	83.69	&	95.06	&	87.35	&	88.8	&	84.8	\\	\hline
\end{tabu}
}
\end{center}
\caption{The table shows DG results on PACS and the effect of various modeling choices. The compiled results compare accuracies across the two optimizing algorithms, the different backbone models, and with and without augmentation.}
\label{tab:comparing_optimizers}
\end{table*}

\subsection{Exploring effects of different modeling choices}
\paragraph{Effect of optimizer:} Table~\ref{tab:comparing_optimizers} compiles the accuracy of different backbones under SGD and ADAM over all domains in PACS. SGD outperforms ADAM across all the backbones. Also, the performance of ADAM is highly susceptible to the learning rate. For instance, averaged over the four domains, SGD gives 89.00\% accuracy compared to 77.69\% accuracy given by ADAM, using the Inception-Resnet backbone. With a higher learning rate (same as SGD), ADAM gives only 48.44\%. The results show that SGD has a clear advantage over ADAM in the studied scenario. The observation may stem from the fact that fine-tuning a large ImageNet model on a relatively small dataset like PACS is an `overparameterized problem'. Wilson~\etal~\cite{wilson2017marginal} suggests that for simple overparameterized problems, adaptive methods can find drastically different solutions than SGD.

\paragraph{Effect of augmentation:} Table~\ref{tab:comparing_optimizers} compares the accuracy of different backbones trained using SGD, with and without augmentation. We observe that augmentations almost always improve the performance of networks. For instance, with the Inception-Resnet model, the average performance of the model across all four domains with augmentation is 89.00\%, which is higher than the average accuracy without augmentation 86.53\%.

\paragraph{Effect of choice of backbone:} Table~\ref{tab:comparing_optimizers} shows the significance of backbone in DG. Across all domains and irrespective of augmentation and other choices, the Inception-Resnet backbone outperforms all other backbones.  Zhou~\etal~\cite{zhou2021domain_survey} questions the common perception that models that perform on ImageNet will learn domain-generalizable features and hence argues for DG tailored methods. In contrast, we observe that the better performing backbone for DG is the better performing model on the ImageNet IID benchmark and not necessarily the backbone with more parameters. 

\begin{table*}[t]
\scriptsize
\begin{center}
\begin{tabular}{| c | c | c | c | c | c | c | c | }
\hline
\textbf{Algorithms}&\textbf{PACS}&	\textbf{VLCS}&	\textbf{Office-home}&	\textbf{Domain-Net}&\textbf{CMNIST}&	\textbf{RMNIST} &\textbf{Average}\\
\hline
IRM~\cite{IRMarjovsky2019invariant}	& 82.9 &	77.2 &	66.7 &	32.6 & 59.16 & 97.7 &69.37\\ \hline
GRL~\cite{ganin2015unsupervised} &	83.69 &	77.38 &	70.2 &	37.4 &	50.5 &	98.49&69.61\\ \hline
MMD~\cite{MMDli2018domain} & 82.8 &	76.7 &	67.1 &	28.4	 & 73.35 & 98.1&71.07\\ \hline
DANN~\cite{DANNganin2016domain} &	84 &	77.7 &	65.5 &	38.1	 & 73.03 & 89.1&71.23\\ \hline
C-DANN~\cite{C-DANNli2018deep} &	81.7 &	74 &	64.7 &	37.9 & 73.03 & 96.3&71.27\\ \hline
DRO~\cite{DROsagawa2019distributionally}	& 83.1 &	77.5 &	67.1 &	33.4 & 	73.35 & 97.9&72.05\\ \hline
RSC~\cite{RSC} &	84.77 &	78.8&	70.8 &	39.2 &	61.2 &	98.23&72.16\\ \hline   
MLDG~\cite{MLDGli2018learning} &	82.4 &	77.1 &	67.6 &	41.6	 & 71.64 & 98&73.05\\ \hline
Mixup~\cite{MIXUPxu2020adversarial} &	83.7 &	78.6 &	68.2 &	38.7	 & 73.34 & 98.1&73.44\\ \hline
CORAL~\cite{CORALsun2016deep} &	83.6 &	77 &	68.6 &	40.2	 & 73.35 & 98.1&73.47\\ \hline 
ERM-Inc-Resnet & 89.11 & 78.84 & 71.95 & 43.2 & 74.35 & 99.2 & 76.10 \\ \hline 
\end{tabular}
\end{center}
\caption{Comparing ERM-Inc-Resnet with other algorithms in DomainBed. The algorithms are sorted by their average performance across the six datasets.}
\label{tab:DG_domainbed}
\end{table*}

\subsection{Comparing with baselines in TDG}
In Table~\ref{tab:DG_domainbed}, we compare the ERM baseline against other algorithms in DomainBed~\cite{gulrajani2020search}. The proposed ERM baseline with Inception-Resnet backbone (ERM-Inc-Resnet) not only outperforms other methods on an average but also outperforms the best performing model in every dataset. On the PACS dataset, we get a margin of above 5\% from the next best performing model. This comparison shows that neural networks trained with a robust backbone, augmentation, and optimizer under TDG setting do not need any additional method to learn domain agnostic features. In supplementary material, we present results by applying the DG tailored methods on top of an optimally trained ERM baseline with the Inception-Resnet backbone. We find that all the studied methods fail to give any improvements over the ERM baseline. In fact, some of the methods simply inhibit learning, reducing test performance. This motivates us to think if, by solving deficits of neural networks in TDG setting, are we attempting to fix a system that is not broken? 


\section{The role of Classwise Priors}
\label{sec:Method}

\begin{figure}[t]
  \centering
  \includegraphics[width=\linewidth]{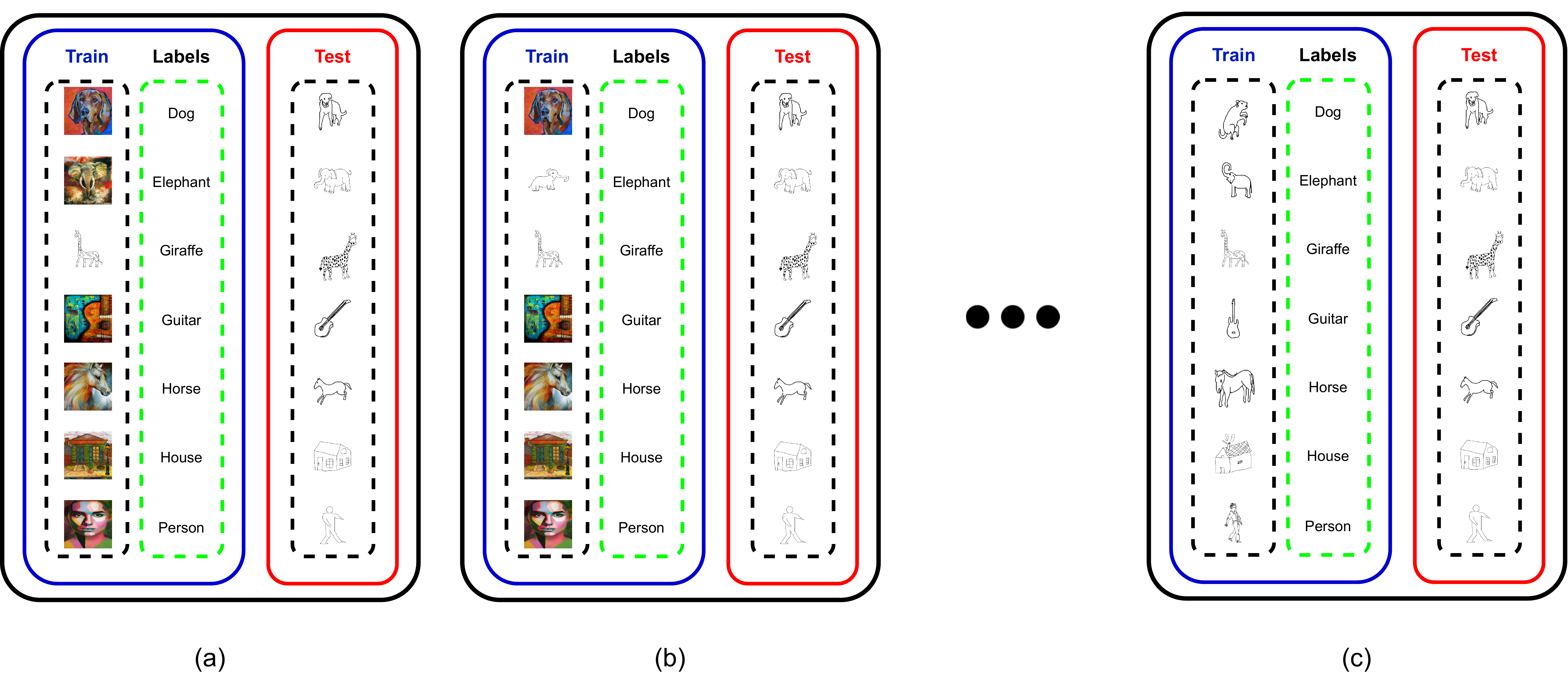}
  \caption{(a) A synthetic data setting, where a domain shift is created on a specific class: `Giraffe'. All train samples are chosen from the domain `Art-Painting' except the class `Giraffe', which is sampled from `Sketch'. At inference, `Sketch' images from across classes are predicted as `Giraffe'. (b)-(c) increasing the number of classes sampled from the sketch domain in the train set.  }
  \label{fig:synth}
\end{figure}

In Section~\ref{domain_aug}, we made an empirical observation that the dataset shifts can explain the performance of an ERM in TDG. In this section, we move beyond TDG and demonstrate that dataset shifts alone cannot explain OOD generalization and the class-wise distribution shifts pose a challenge. To make our case, we formulate a synthetic setting using two domains (`Art-Painting' and Sketch') from the PACS dataset. For the first model, the training data is created with a domain shift on a specific class. All train samples are chosen from the domain `Art-Painting' except the class `Giraffe,' which is sampled from `Sketch' (as illustrated in Figure~\ref{fig:synth}). We then train a second network using all classes from a single domain (`Art-Painting'). Both models are evaluated on the `Sketch' domain. We use Frechet Inception Distance (FID)~\cite{heusel2017gans} to measure the dataset shift among train-test splits in the two settings. 

When the training data for all the classes are picked from a single domain (`Art-Painting'), we get a test performance of 61.67\% on the test data (`Sketch'). By replacing a single class (`Giraffe') in training from the `Sketch' domain, all test images are predicted as Giraffe. The accuracy drops to 13\%, the fraction of samples from the class Giraffe. The observation is intuitive as the network fits the easier discriminative variances that differentiate domains and use them to classify.  

 Unlike TDG setup, this behavior cannot be explained by distribution shifts among the train and test split. Post adding the 'Giraffe' class; the train data is now closer to test data as they share a domain for a class. The FID score of train and test data in the first setting is 147.50, and in the second setting is 95.96. The performance drops despite the reduction in the dataset shift.

The above observation shows that there are factors other than distribution shifts that determine the OOD generalization of neural networks. The tendency of a network to fit the domain-specific variances which are used to distinguish classes in train data is caused by the priors in train data (different domains see a different subset of classes). The model's ability to escape fitting such prior can be evaluated by extending this synthetic experiment, which precisely motivates the need for CWDG evaluation. In the CWDG setting, different seeds for sampling domains from classes to create test data can evaluate the different variances that a model fits.  

We increase the proportion of `Sketch' domain in train data (Figure~\ref{fig:synth} (b)-(c)), adding one class at a time: `giraffe', `elephant', `dog', `guitar', `horse', and `house'. The corresponding percentage accuracies are 13.3, 55.7, 73.88, 75.8, 96.02, and 96.5, respectively. For example, having both Elephant and Giraffe from the Sketch domain in training increases the test accuracy on the Sketch domain to 55.7\%. The domain features become less useful to discriminate classes in train data; hence the effective prior reduces.

\subsection{Classwise Domain Generalization}

We follow the notations presented in \cite{PACS}. We consider $S$ available domains containing samples from a total of $C$ classes. Let $c_i$ represent the class $c$ corresponding to $i^{th}$ domain. Furthermore, let $N_{c_i}$ be the number of labeled samples in class $c$ and domain $i$. $y_{c_i}(k)$ represents the label corresponding to $k^{th}$ such sample and $\hat{y}_{c_i}(k)$, the label prediction. $\ell$ is the loss computed between the true and predicted labels. 
    The objective for minimization in the TDG setting is as follows:

\small$$\underset{\Theta_{}}{\operatorname{argmin}} \frac{1}{S-1} \sum_{i=1 : S_i \neq S_{test}}^{S} \frac{1}{C} \sum_{(c_{i} \in C)} \frac{1}{N_{c_i}} \sum_{k=1}^{N_{c_i}} \ell\left(\hat{y}_{c_i}(k), y_{c_i}(k)\right),$$ \normalsize
where $\Theta_{}$ are the network parameters. 

In the proposed CWDG setting, the optimization problem is modified as follows:

\small $$\underset{\Theta_{}}{\operatorname{argmin}} \frac{1}{S} \sum_{i=1}^{S} \frac{1}{C_{S_{i}}} \sum_{\left( c_{i} \in C: c_i \neq c_{i_{test}}\right)} \frac{1}{N_{c_i}} \sum_{k=1}^{N_{c_i}} \ell\left(\hat{y}_{c_i}(k), y_{c_i}(k)\right).$$
\normalsize

Where, $C_{S_{i}}$ is the number of classes present in the domain $S_{i}$ for train. 
Assume $\Theta_{*_1}$ is obtained post minimization in TDG setting, and $\Theta_{*_2}$ is the solution obtained under CWDG. $\Theta_{*_1}$ is evaluated on unseen domain $S_{test}$. On the other hand, $\Theta_{*_2}$, which has been optimized over all domains in $S$, is evaluated on unseen class instances $c_{i_{test}}$. The solution in CWDG, $\Theta_{*_2}$, intuitively needs to be able to discriminate classes agnostic to which domain it appears in $S$, but our empirical evidence show otherwise.

We look at the type of shift incurred in CWDG compared to classical TDG setting following the study of datashift by Moreno-Torres~\etal~\cite{moreno2012unifying}. They study various data shift and broadly classify them into four different categories, namely: 
\begin{itemize}
    \item Covariate shift: $P_{test}(x) \neq P_{train}(x)$ but $P_{test}(y|x) = P_{train}(y|x)$
    \item Prior probability shift: $P_{test}(y) \neq P_{train}(y)$ but $P_{test}(y|x) = P_{train}(y|x)$
    \item Concept shift : $P_{test}(x) = P_{train}(x)$ but $P_{test}(y|x) \neq P_{train}(y|x)$
    \item Dataset shift: $P_{test}(x,y) \neq P_{train}(x,y)$ but none of the above hold.
\end{itemize}

CWDG falls under the category of dataset shift, while TDG formulation falls under covariate shift. Both measures of robustness are distinct and desirable for corresponding real-world applications. It is worth noting that TDG is a special case of CWDG, where the same domain is kept out from each class.

\subsection{Iterative Domain Feature Masking}
We propose a method to iteratively mask the features which contribute significantly for domain classification. We augment the network with an additional branch to predict the domain. The two branches are trained in parallel, one predicting domain and another predicting class for a given sample. In the two-branch network, the shared feature at the branched layer (the last layer) $g$ is given by $g = f(\theta, x)$ where $x$ is the input image and $\theta$ are the parameters of the network in all but the last layer. The class predictions and domain predictions are given by $y_{class} = f_1(\theta_1, g)$ and $y_{domain} = f_2(\theta_2, g)$. $\theta_1$ and $\theta_2$ are the weights of the class and domain prediction layers, respectively. 

We compute the gradient of the predicted domain with respect to $g$, as  $\ grad = \partial(f_2(g; \theta_2)\cdot \hat{y}_{domain})/\partial g $, given the ground truth domain label $\hat{y}_{domain}$. IDFM computes a threshold $q_{th}$ such that the value of top $q$ percentile of elements in $|grad|$ is above $q_{th}$. $q$ is a hyper-parameter which we choose as 33\% throughout our experiments. A mask M is computed corresponding to the $i^{th}$ element in $g$ such that,


\begin{equation}
  M =
    \begin{cases}
      0 & \text{if $grad(i) \geq q_{th}$}\\
      1 & \text{otherwise}
      \end{cases}       
\end{equation}

Now we compute the masked feature $g'$ as: $\ g' = g \cdot M$. The new class prediction, post masking the dominant features corresponding to domain information is  $y_{class} = f_1(\theta_1, g')$. This is done iteratively in every step so that the class prediction is not reliant on domain-specific features.


The proposed IDFM method is inspired from the two previous methods, i.e., Representative Self Challenging (RSC)~\cite{RSC} and gradient reversal~\cite{ganin2015unsupervised}. RSC is a single branch network that iteratively masks the features with the highest say (gradient) in the class prediction. The assumption here is that the dominant features activated on train data are domain-specific. We explicitly compute the features that contribute the most to domain prediction and mask them, freeing the method of the aforementioned assumption. The gradient reversal (GRL) algorithm manipulates features through gradients flowing from the domain prediction branch to make them domain agnostic. In contrast, we explicitly mask domain-specific features in the last layer.



\subsection{CWDG Benchmark}

We compare the performance of eleven DG algorithms on six different datasets against the proposed IDFM method to comprise the CWDG benchmark. This includes our implementations of ERM-Inc-ResNet and GRL. All the methods are trained on top of the Inc-ResNet backbone. We present results on one of the possible splits (randomly selecting a test domain for each class). Results on a few other splits are presented in the supplementary material. We use the same augmentations and hyper-parameters as in DomainBed.



\begin{table*}[t]
\scriptsize
\begin{center}
\begin{tabular}{| c | c | c | c | c | c | c | c | }
\hline
\textbf{Algorithms}&\textbf{PACS}&	\textbf{VLCS}&	\textbf{Office-home}&	\textbf{Domain-Net}&\textbf{CMNIST}&	\textbf{RMNIST} & \textbf{Average}\\
\hline

IRM~\cite{IRMarjovsky2019invariant}	& 64.8 & 63.1 &	55.77 &	28.8 & 61.58 & 71.2 & 57.53 \\ \hline
RSC~\cite{RSC} &	79.3 &	64.5 &	65.2 &	25.3 &	50.5 &	98.7 &  63.91 \\ \hline
MMD~\cite{MMDli2018domain} & 73.8 &	60.2 &	65.46 &	25.8	 & 73.05 & 98.5 & 66.13 \\ \hline
DANN~\cite{DANNganin2016domain} &	74.4 &	64.2 &	62.95 &	24.6	 & 72.05 & 98.8 & 66.16 \\ \hline
MLDG~\cite{MLDGli2018learning} &	73 &	62.97 &	65.87 &	25.73	 & 71.93 & 98.5 & 66.3 \\ \hline
CORAL~\cite{CORALsun2016deep} &	77.06 &	60.2 &	65.46 &	25.8	 & 73.5 & 98.5 & 66.75 \\ \hline
C-DANN~\cite{C-DANNli2018deep} &	77.7 &	63.77 &	64.58 &	24.04 & 72.5 & 98.9 & 66.91 \\ \hline
ERM-Inc-Resnet & 79.6 & 60.86 & 66.1 & 25.8 & 71.15 & 99.8 & 67.21 \\ \hline 
Mixup~\cite{MIXUPxu2020adversarial} &	77.6 &	64.2 &	66.02 &	25.1	 & 73.05 & 98.3 & 67.35 \\ \hline
DRO~\cite{DROsagawa2019distributionally}	& 79.38 & 64.77 & 66.1 &	25.15 & 73.05 & 98.5  &  67.82\\ \hline
GRL~\cite{ganin2015unsupervised} & 86.2 & 65.2 & 66.9 & 26.1 & 74.15 & 99.3 & 69.64 \\ \hline
IDFM (ours) & 88.84 & 67.87 & 66.9 & 26.9 & 75.32 & 99.7 & 70.92 \\ 
\hline
\end{tabular}
\end{center}
\caption{Comparing the performance of different algorithms in DomainBed against Inception-Resnet-ERM and GRL(with Inception-Resnet backbone) in CWDG setting. Algorithms are sorted by their average performance across the six datasets.}
\label{Tab:classwise-dg}
\end{table*}


Table \ref{Tab:classwise-dg} illustrates the obtained results. A one-to-one comparison between results in TDG and CWDG settings is not entirely meaningful. However, it is worth noting that despite seeing all domains during training, the obtained accuracies are lower than their TDG counterpart. The comparison suggests that classwise priors pose a significant challenge in OOD generalization. 


We further observe that the idea of gradient reversal (GRL) holds merit in CWDG formulation giving notable improvements over the existing DG methods. It is worth noting that GRL deteriorated performance in TDG, further motivating the need to evaluate DG from varied perspectives. IDFM achieves state-of-the-art results, demonstrating the efficacy of the proposed approach. The results also indicate that explicit feature masking seems to improve over gradient-based feature manipulation in the studied setting, and the idea may be worth exploring in alternate settings like domain adaptation. 



\section{Conclusion}
\label{sec:Conclusion}
In recent years, the efforts in Domain Generalization have focused on proposing tailored methods in TDG. In contrast, our work focuses on fundamental analysis around the problem. It extends and explains some of the findings in~\cite{gulrajani2020search}. We show that in TDG, a carefully trained ERM is not just competent; it clearly outperforms all existing state-of-the-art methods on six standard datasets. We present empirical evidence that it is favorable for an ERM to move towards domain agnostic features in the presence of multiple domains (classwise equally distributed). We further uncover the role of priors and present the CWDG benchmark. Unlike TDG, in the CWDG framework, the network tends to learn domain-specific features. The observation is confirmed by performance improvements using explicit methods for suppressing domain-specific features. The proposed IDFM approach achieves SOTA results on the CWDG benchmark and provides an improved variation of the seminal gradient reversal method. We believe the presented reappraisal of DG furthers the understanding of OOD generalization and will aid future efforts. 

\bibliographystyle{splncs04}
\bibliography{mybibliography}

\clearpage
\appendix

\title{Appendix}
\maketitle

\section{Introduction}
\label{sec:intro}
In appendix, we present four sections. First section details the training strategy followed for comparing baselines in both TDG and CWDG baseline comparison experiments. The second section shows the effect of SOTA methods in DomainBed on Inception-Resnet backbone. In the subsequent section, we show the results of different seeds for train test split in CWDG. In the last section we discuss the challenge in model selection in TDG. 

\section{Training Details}

For augmentation, we use standard image perturbations as used in DomainBed \cite{gulrajani2020search}. We select random crops from train images and resize them into model input size. We randomly flip half the images in every batch horizontally (with 0.5 probability) and use color-jitter augmentation by randomly scaling brightness, contrast, saturation, and hue. These scales are sampled randomly from a distribution between 0.6 and 1.4. We randomly choose images from each batch and grayscale them (with a probability of 0.1). We normalize all images with ImageNet mean and standard deviation. We use the hyperparameters and random-seed used in DomainBed to maintain uniformity of comparison. We follow the augmentation strategy used in DomainBed. This follows that MNIST datasets, namely CMNIST~\cite{IRMarjovsky2019invariant} and RMNIST~\cite{rmnistghifary2015domain} are not augmented.

We train each model for 30 epochs with a batch size of 32. We use a learning rate of 0.01 with no scheduler across all the runs, except for Alexnet (learning rate of 0.001) since it does not converge at higher values. We use ImageNet pre-trained weights for all the backbone models and fine-tune the last layer with a categorical-cross-entropy loss function. We use a weight decay of 0.0005 and momentum of 0.9.

\section{Effect of SOTA methods on Inception-Resnet Backbone}
\label{sec:sec1}

In this section we present the results of evaluating the efficacy of different SOTA TDG methods using Inception-Resnet~\cite{inception_resnet} backbone on five different datasets~\cite{PACS,fang2013unbiased,officehomevenkateswara2017deep,IRMarjovsky2019invariant,rmnistghifary2015domain}. We Follow the training strategy in DomainBed~\cite{gulrajani2020search} similar to the experiment in Section 4.2 in main text. 

Table~\ref{tab:DG_domainbed_inception} shows the performance of the ten methods namely:\\ \cite{C-DANNli2018deep,IRMarjovsky2019invariant,MLDGli2018learning,DROsagawa2019distributionally,MMDli2018domain,gulrajani2020search,CORALsun2016deep,MIXUPxu2020adversarial,RSC,DANNganin2016domain,ganin2015unsupervised} with Inception Resnet backbone on the five different datasets. We can see that despite having the best performing backbone, none of the methods outperform ERM.

\begin{table*}[t]
\footnotesize
\begin{center}
\begin{tabular}{| c | c | c | c | c | c | c |}
\hline
\textbf{Algorithms}&\textbf{PACS}&	\textbf{VLCS}&	\textbf{Office-home} &\textbf{CMNIST}&	\textbf{RMNIST} &\textbf{Average}\\
\hline

IRM	& 88.9 &	75.8 &	56.6 &	61.54 & 79.2 & 72.40 \\ \hline
GRL &	86.4 &	75.4 &	65.8 &	51.6 & 98.49 &	75.53\\ \hline
MLDG &	76.4 &	72.6 &	67.6 &	73.4	 & 98.1 & 77.62\\ \hline
RSC &	85.87 &	77.8&	70.8 &	62.5 &	98.23 &	79.04\\ \hline   
C-DANN &	85.7 &	75.8 &	66.78 &	74.25 & 97.95 & 80.9\\ \hline
MMD & 88.3 &	78.2 &	67.8 &	73.89	 & 98.3 & 81.3\\ \hline
CORAL &	87.5 &	77.8 &	68.89 &	74.25	 & 98.2 & 81.32 \\ \hline
DRO	& 88.9 &	78.5 &	66.7 &	73.89 & 	98.2 & 81.34\\ \hline
Mixup &	88.8 &	78.7 &	67.8 &	74.2	 & 97.89 & 81.48\\ \hline
DANN &	89.0 &	78.7 &	68.8 &	74.25	 & 97.8 & 81.77\\ \hline
ERM & 89.11 & 78.84 & 71.95 & 74.35 & 99.2 & 82.7 \\ \hline 
\end{tabular}
\end{center}

\caption{Comparing ERM-Inc-Resnet with other methods in DomainBed implemented with inception-Resnet backbone. The algorithms are sorted by their average performance across the five datasets.}
\label{tab:DG_domainbed_inception}
\end{table*}

\section{Different seeds for CWDG}
\label{sec:sec2}
In this section we present results for the different train test splits in CWDG setting of the PACS dataset. We observe comparable performance across the different seeds of CWDG(Table~\ref{tab:CWDG_seed}). Each column in the table correspond to one run and the last row in each column shows the accuracy of ERM Inc-ResNet on that run. That is, each row correspond to a class and each element shows the domain kept out for the particular class. The first column shows the split used in the main text. The accuracies shows that as long as the domains are evenly spread such that there is a clear prior in the train split the performance of neural network stays signficantly below the TDG setting. 

\begin{table}[t]
\footnotesize
\begin{center}
\begin{tabular}{| c | c | c | c | c | c |}
\hline
\textbf{Classes}&\textbf{Domain}&	\textbf{Domain}&	\textbf{Domain} &\textbf{Domain} &\textbf{Domain} \\
\hline

\textbf{Guitar} &	Photo  &	Art &	Cartoon &	Sketch &	Art\\ \hline
\textbf{Person} &	Photo &	Cartoon &	Art &	Photo &	Cartoon\\ \hline
\textbf{Horse} & 	Cartoon &	Sketch 	& Cartoon & Art & 	Photo\\ \hline
\textbf{Elephant} &	Sketch &	Photo &	Art &	Sketch &	Art\\ \hline
\textbf{Dog} &	Photo &	Art & Sketch	& Cartoon & Cartoon\\ \hline
\textbf{Giraffe} &	Art &	Photo & Cartoon	& Art &	Photo\\ \hline
\textbf{House} &	Cartoon &	Cartoon & Photo	& Sketch & Sketch\\ \hline
Accuracy &	79.6 &	79.38 &  78.81	& 79.12 & 79.18\\ \hline
\end{tabular}
\end{center}
\caption{performance of NN on different train test splits in CWDG setting. Last row of each column shows results of one run and each row of the column corresponds to the domain kept out for the corresponding class.}
\label{tab:CWDG_seed}
\end{table}

\begin{table}[t]
\footnotesize
\begin{center}
\begin{tabular}{| c | c | c | c | c |}
\hline
\textbf{Train domains}&\textbf{Test domain}&	\textbf{$A_{min}$}&	\textbf{$A_{max}$} &\textbf{$A_{sel}$} \\
\hline

SAC &	P &	94.01 &	96.4 &	96.3\\ \hline
PSC &	A &	78.7 &	86.3 &	84\\ \hline
PAS &	C &	81.1 &	87.8 &	83\\ \hline
PAC &	S &	64.96 & 85.09	& 83.2 \\ \hline

\end{tabular}
\end{center}
\caption{Model selection in TDG}
\label{tab:model_sel}
\end{table}

\section{Model selection in DG}
\label{sec:sec3}
In the TDG setting, model selection is traditionally done using the test data (the entirety of the unseen domain). However, the formulation of DG implies that the test domain is not known during training. Hence, using an oracle for model selection is not appropriate in the strict sense. To further analyze the challenge in model selection in the TDG setting, we contrast the best and worst models (based on their test performance), picked after the training saturates (the loss curve saturates). 

We run Inception-Resnet-ERM~\cite{inception_resnet} in the TDG setting of the PACS~\cite{PACS} dataset. For each run, we keep out one domain for test. We also create a validation set by randomly sampling 5\% data from each of the three training domains and monitor the model performance on this validation set after each training step. We select the best-performing model on this validation set. We report the accuracy of this model on the test domain ($A_{sel}$). We also report the performance of the model selected by an oracle for each run as $A_{max}$ (directly using test data for validation). $A_{min}$ is the accuracy of the model that gives the lowest test performance once the training accuracy saturates. We report $A_{max}$, $A_{min}$ and $A_{sel}$ with all four folds of PACS data.

From Table~\ref{tab:model_sel} we can see that model selection in the TDG setting is not trivial. Performance on a validation set (sampled from train split) does not guarantee performance on the unseen domain. To evaluate the possible model selection for the CWDG setting, we run Inception-Resnet-ERM in the CWDG setting of the PACS dataset. The $A_{sel}$, $A_{max}$ and $A_{min}$ value for the CWDG setting are 79.1\%, 79.4\% and 66.3\% respectively. The difference between the $A_{max}$ and $A_{min}$ in the CWDG setting is higher than the average difference in the TDG setting. The model picked using the validation set gives comparable performance to the oracle.

\end{document}